# Anomaly Detection in Video Using Predictive Convolutional Long Short-Term Memory Networks

Jefferson Ryan Medel    Andreas Savakis

Rochester Institute of Technology, Rochester, New York
andreas.savakis@rit.edu

## Abstract

*Automating the detection of anomalous events within long video sequences is challenging due to the ambiguity of how such events are defined. We approach the problem by learning generative models that can identify anomalies in videos using limited supervision. We propose end-to-end trainable composite Convolutional Long Short-Term Memory (Conv-LSTM) networks that are able to predict the evolution of a video sequence from a small number of input frames. Regularity scores are derived from the reconstruction errors of a set of predictions with abnormal video sequences yielding lower regularity scores as they diverge further from the actual sequence over time. The models utilize a composite structure and examine the effects of 'conditioning' in learning more meaningful representations. The best model is chosen based on the reconstruction and prediction accuracies. The Conv-LSTM models are evaluated both qualitatively and quantitatively, demonstrating competitive results on anomaly detection datasets. Conv-LSTM units are shown to be an effective tool for modeling and predicting video sequences.*

## 1. Introduction

Anomalies in videos are broadly defined as events that are unusual and signify irregular behavior. Consequently, anomaly detection has broad applications in many different areas, including surveillance, intrusion detection, health monitoring, and event detection. Unusual events of interest in long video sequences, e.g. surveillance footage, often have an extremely low probability of occurring. As such, manually detecting these rare events, or anomalies, is a very meticulous task that often requires more manpower than is generally available. This has prompted the need for automated detection and segmentation of sequences of interest [1]-[15].

In contrast to the related field of action recognition where events of interest that are clearly defined, anomalies in videos are vaguely defined and may cover a wide range of activities. Since it is less clear-cut, models that can be trained using little to no supervision, including spatio-temporal features, dictionary learning and autoencoders [15] have been applied to evaluating anomalies.

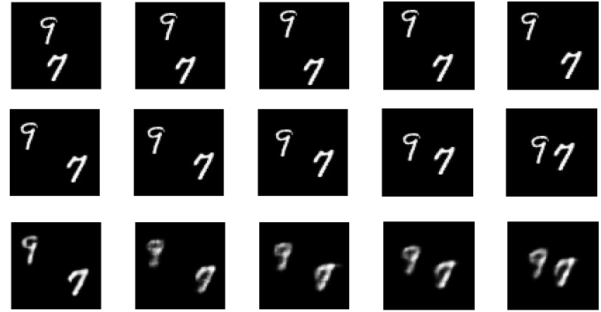

Figure 1. Future prediction example on the bouncing MNIST dataset. Columns indicate time steps. Top row: input sequence; Middle row: Ground truth for next 5 frames; Bottom Row: Our composite Conv-LSTM Network prediction of the next 5 frames.

This paper makes two main contributions. The first contribution is the development of a generative model, based on a composite Convolutional Long Short-Term Memory (Conv-LSTM) network architecture. Inspired by [16], our Conv-LSTM network incorporates a composite model that is able to encode an input video sequence, reconstruct it, and predict its near term future. An example of the predictive capability of our network in shown in Fig. 1. Our model utilizes Conv-LSTM units that allow the network to better learn spatio-temporal representations. We extend the model of [16] by creating a composite model with better predictive power and considering both an unconditioned and a conditioned variant where the output is used to condition the input of the next timestep.

Our second contribution is the application of the Conv-LSTM network to detect anomalous video segments using a regularity evaluation algorithm at the model's output. The regularity of a video sequence is relative to other sequences from the same source. The models are evaluated on the UCSD Pedestrian 1 dataset, UCSD Pedestrian 2 dataset, Subway dataset, and Avenue dataset.

## 2. Related Works

Anomaly detection may be viewed as binary classification problem when ground truth labels are available for anomalous actions. However, such labels are often uncommon or unwieldy, and the data available for training a model is limited to containing little to no anomalous events. The available training data often results in the formulation of semi-supervised models that can be adapted to operate in an unsupervised mode by using a sample of the unlabeled data set as training data. Such techniques have been used to great effect in [1]-[15], where models are trained with little to no supervision and used to classify anomalous sequences in a given video.

Zhao et al. utilizes an unsupervised dynamic sparse coding algorithm in [9] to train dictionaries with which anomalies are detected through the reconstruction error. Lu et al. improves upon this in [14] by introducing an approach that directly learns sparse combinations instead of a dictionary, thereby significantly speeding up testing. While sparse coding has been shown to be effective, dictionaries may still contain unused or noisy elements within the dictionary, reducing their effectiveness.

Handcrafted features comprised of a mixture of dynamic textures and spatial anomaly maps are used by Cong et al. in [10] to learn the "normalcy" of a video sequence, with which an anomaly score is computed. Similarly, Adam et al. [3] created a probability distribution of low-level observations. Given a new observation, the likelihood of occurrence is used to determine whether or not it is anomalous. A limitation of hand-crafted features is that they may be unable to adapt to or learn unexpected features effectively. Deep learning addresses this issue by allowing the network to learn what features are important.

Hasan et al. employs a convolutional neural network (CNN) in [15] to learn the temporal regularity of given video sequences. A regularity score is computed from the reconstruction error and used to detect anomalous segments. While effective, CNNs were not developed with temporal features in mind and are not a natural fit for video. Conv-LSTMs generalize CNNs in a manner similar to how LSTMs generalize dense networks.

## 2.1. Convolutional LSTM Units

A Convolutional Long Short-term Memory architecture was recently utilized by Shi et al. in [16] and Patraucean et al. in [17]. The Conv-LSTM units integrate convolution into the fully connected LSTM (FC-LSTM) by replacing the weights with convolutional filters. The formulation of the Conv-LSTM unit is summarized in equations (1) to (5) and illustrated in Figure 2(a), where the convolutions are carried out at the weighted connections.

$$I = \sigma(W_{XI} * X_t + W_{HI} * H_{t-1} + W_{CI} \circ C_{t-1} + b_I) \quad (1)$$
$$F_t = \sigma(W_{XF} * X_t + W_{HF} * H_{t-1} + W_{CF} \circ C_{t-1} + b_F) \quad (2)$$
$$C_t = F \cdot C + i_t \circ (W_{XC} * x_t + W_{HC} * h_{t-1} + b_C) \quad (3)$$

$$O_t = \sigma(W_{XO} * X_t + W_{HO} * H_{t-1} + W_{CO} \cdot C_{t-1} + b_o) \quad (4)$$
$$H_t = O \circ \tanh(C_t) \quad (5)$$

The input, forget, cell, output, and hidden state of each timestep are denoted by $I$, $F$, $C$, $O$, and $H$ respectively, the activation by $\sigma$, and the weighted connections between states by a set of weights, $W$. The output state controls how much information is propagated from the previous timestep, while the hidden state consists of information received by the next timestep and layer. The peephole connections allow the LSTM unit to access and propagate information recorded from the cell state of the previous timestep.

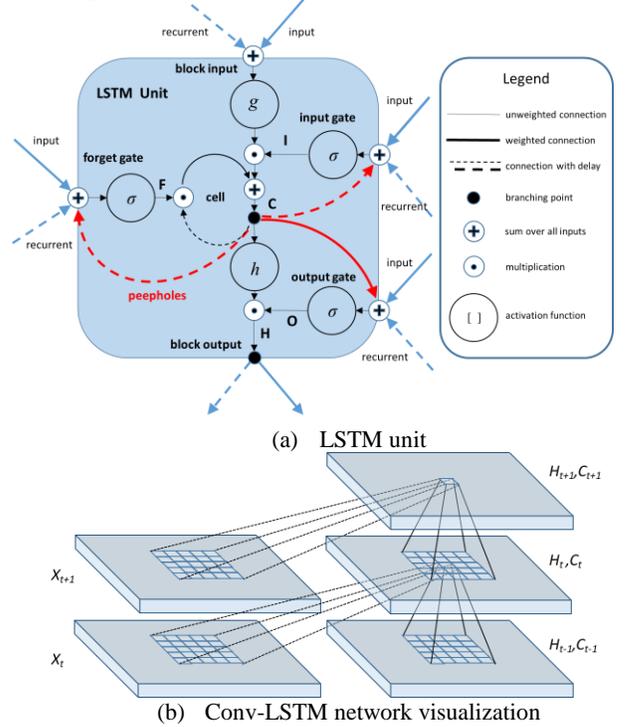

(a) LSTM unit

(b) Conv-LSTM network visualization

Figure 2. Conv-LSTM network operation (a) LSTM unit; (b) Conv-LSTM network visualization.

The Conv-LSTM network is more advantageous than the FC-LSTM when working with images, due to its ability to propagate spatial characteristics temporally through each Conv-LSTM state. The FC-LSTM may be viewed as a special case of Conv-LSTM, where the filter size is equal to the input image and a single convolutional operation is performed, such that each Conv-LSTM unit shares the same parameters through all timesteps.

Just as the convolutional filters of the input-to-hidden connections determine the resolution of feature maps created from the input, the convolutional filter size of the hidden-to-hidden connections determines the aggregate information the Conv-LSTM unit receives from the previous timestep. A visualization of the process is shown in Figure 2(b). The transition of states between timesteps for a Conv-LSTM unit can then be interpreted as movement between frames. Larger transitional kernels tend to capture faster motions while smaller transitional kernels capture slower motions, as in [16].

## 2.2. Future Video Prediction

Long Short-Term Memory networks are capable of learning long-term dependencies. As such, they are able to extrapolate temporally sequential data given certain inputs. Srivastava et al. [18] take advantage of this property to train a composite encoder-decoder model able to reconstruct the past and predict the future video sequences. More specifically, the encoder maps an input video representation to a fixed length representation, while the decoder extrapolates the learned encoding into the past and future video sequences.

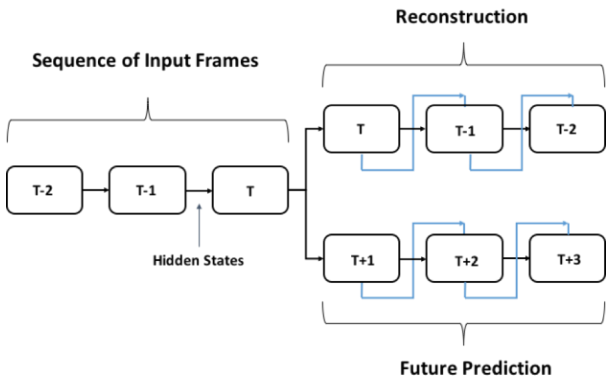

Figure 3. The composite structure for unrolled LSTM unit. Blue lines represent potential conditioning.

When using a simple encoder-decoder model, the target values of the model determine the model application. When the target output is the input, the model creates a reconstruction of the input video sequence. When the target output is subsequent frames, the model learns to predict the future of the video sequence. The simple encoder-decoder model is improved by combining both the reconstruction and prediction models into a composite model, as seen in Fig. 3. Both the current and future video sequences are target outputs. Reconstruction models tend to learn trivial representations that merely memorize the input. Future predictors tend to absorb more information from the most recent frames, as they are generally the most immediately relevant. While this is effective for specific predictions, the loss of information from older timesteps can lead to less accurate predictions for more general video sequences. The reconstruction of both the past and future video sequences forces the learned encoding to contain more meaningful data, thus improving the overall performance of the system.

Shi et al. propose the use of Convolutional LSTMs with the encoder-decoder structure to better retain spatio-temporal information [16]. Its decoder is unique in that it performs a 1x1 convolutional operation across the output of each layer to obtain an output, as opposed to looking solely at the last layer. The proposed architecture was shown to outperform the LSTM models used by [18] when predicting future video sequences for a synthetic Bouncing MNIST Dataset. It was also successfully applied to a precipitation forecasting problem that used satellite imagery of clouds to predict weather patterns.

## 3. Anomaly Detection with Conv-LSTMs

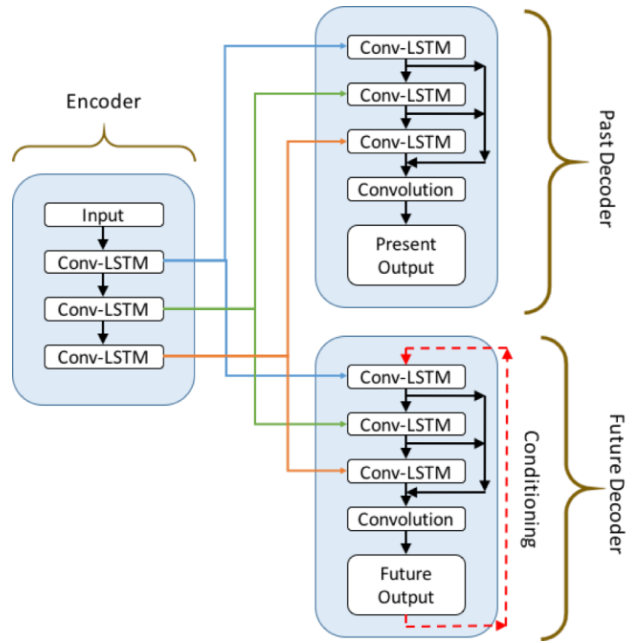

Figure 4. High-level view of the Conditioned Composite Conv-LSTM Encoder-Decoder.

## 3.1. Composite Conv-LSTM Encoder-Decoder

We propose a Composite Conv-LSTM Encoder-Decoder to learn the regularity of videos from non-overlapping patches of frames from an input segment. The limitation in area forces the architecture to produce a more meaningful encoding. The network learns to accurately predict 'normal' actions similar to those found in the videos used to train it. This causes the prediction of abnormalities to diverge further from the ground truth with each subsequent timestep. The regularity scores derived from the reconstruction errors can therefore be used to determine when anomalies occur within videos.

A high level view of the proposed model using three Conv-LSTM layers is shown in Fig. 4. The proposed architecture utilizes multiple stacked Conv-LSTM layers in an end-to-end trainable network. The design has two main elements, the encoder and the decoder.

**Encoder** The encoder accepts as input a sequence of reshaped frames in chronological order. By reshaping the input into a stack of non-overlapping patches, the model loses some detail but learns the more significant characteristics of the data. Each Conv-LSTM layer is made up of multiple Conv-LSTM units that span across the specified number of timesteps. The outputs of the last timestep of each Conv-LSTM layer are used as the encoding. Unlike traditional convolutional neural networks [20], the proposed model does not utilize max-pooling layers. It instead feeds the output of each Conv-LSTM layer directly into the next one.

**Decoder** There are two decoders, one reconstructing the past input video sequence and the other predicting the future. Each decoder is initialized with the encoding provided by corresponding layers in the encoder, as shown in Fig. 4. The past decoder output is determined solely from the information extracted from its initialization. The outputs of each layer are concatenated and summed through a 1x1 convolutional filter to obtain the reconstruction of the input.

We consider two options for the future decoder: unconditioned and conditioned. An unconditioned decoder has the same architecture as the past decoder. A conditioned decoder uses the summed output of each timestep as the input to the first layer of the subsequent timestep, thus 'conditioning' it to the previous frame.

As in [18], only the future decoder is conditioned, because the past has one possible outcome, while the future may vary. Conditioning potentially limits the variation by providing more information from the previous timestep. The proposed architecture uses a composite conditioned structure comprised of Conv-LSTM units. This potentially allows the model to better learn a video's normality, thus making anomalous video segments containing sequences more likely to stand out because they are hard to reconstruct.

### 3.2. Anomaly Evaluation Algorithm

A trained model can be used to obtain a reconstruction of the input video sequence and predict its future frames. In a qualitative inspection of the reconstruction and predictions, anomalous events are more likely to stand out, as the trained model does not have the necessary information to accurately reconstruct or predict anomalies. The reconstruction error is expressed as the Mean Square Error (MSE) defined in eq. (6), where $\hat{\theta}$ is the output pixel value, $\theta$ is the target value, $p$ is the total number of pixels per frame, and $n$ is the number of frames.

$$e = \frac{1}{np}\sum_{k=1}^{n}\sum_{i=1}^{p}(\hat{\theta}_{ki} - \theta_{ki})^2 \qquad (6)$$

The quantitative evaluation algorithm considered here is based on a regularity score that is computed from the error values [15]. The regularity score normalizes the error of the reconstruction between zero and one with respect to the other reconstructions from the same video, as different videos may have different notions of abnormality. The regularity $g(x)$ of a sequence can be computed as follows:

$$g(x) = 1 - \frac{e(x) - \min_X e(x)}{\max_X e(x)} \qquad (7)$$

where $x$ is the output reconstruction sequence and $e(x)$ is the reconstruction error of that sequence. Video sequences containing normal events have a higher regularity score since they are similar to the data used to train the model, while sequences containing abnormal events have a lower regularity score. Distinct local minima or scores below a certain threshold from a time series of regularity scores can therefore be used to locate abnormal events.

We utilize of both distinct local minima and maxima found using the Persistence 1D algorithm [21]. Distinct local minima represent video frames that are highly likely to contain anomalies. Regions of anomalous video segments are proposed based on the minima found. Points that are within a certain threshold of one another are considered to be part of the same anomalous sequence.

Distinct local maxima potentially represent regular video sequences that take place immediately before or after an anomalous event. These points help decrease the number of false positives by limiting the length of the proposed anomalous segment. More specifically, the new proposed region border is midway between the maxima and minima. This constraint is precluded when the distinct maximum is between two distinct minima that are considered to be of the same anomalous sequence. During evaluation, anomalies are considered detected if a certain percentage of the proposed detection is overlapped by the ground truth anomaly. Multiple detections of the same anomaly are considered a single true positive detection.

## 4. Results

### 4.1. Experimental Setup

**Datasets** The datasets used in the experiments are the UCSD Pedestrian Datasets [8], Avenue Dataset [14], and Subway Datasets [3]. The UCSD Pedestrian dataset consists of two datasets, the UCSD Pedestrian 1, and UCSD Pedestrian 2 datasets, containing video clips of two different walkways. The Subway dataset contains two videos, one of the station's "entrance gate", and one its "exit gate". The datasets were chosen for their variety of anomalous actions.

The proposed models are initially evaluated to determine the most effective. The selected model is then trained and evaluated over multiple datasets. The outputs are analyzed qualitatively through visualization of the past reconstruction and future prediction, and quantitatively by comparing the average loss and detection rates.

## 4.2. Parameter Selection

**Model Parameters** The input images are resized to 224x224 pixels and converted to grayscale. A preliminary Conv-LSTM Encoder-Decoder baseline model was evaluated for use as reference in parameter selection. The baseline model utilizes an input and output length of five, and divides the image into 64 non-overlapping patches. Using the model from [16] as reference, a filter size of 5x5 and three Conv-LSTM layers are used, while the total number of filters is doubled from 256 to 512 to accommodate the larger image size. The Conv-LSTM units use sigmoid nonlinearities for the input, output and forget states, and tanh for the hidden and cell states. A sigmoid non-linearity is applied to the final convolutional layer. Padding is applied during all convolution operations to retain the image size. The baseline model is simpler than the composite one and utilizes only a future decoder.

The parameters tested in variations of the baseline model include the length of both the input and output timesteps, the filter size, and the final output non-linearity function, as shown in Table 1. The model with an input length of ten outputs a lower MSE per frame, but takes 1.5 times longer to train, which makes it largely inefficient. The more timesteps into the future the model predicts, the worse each prediction becomes. A filter size of 3x3 was considered for capturing smaller motions, but was not as effective. The commonly used rectified linear unit (ReLU) nonlinearity function was tested at the final output, but ultimately, the parameters used by the baseline model were found to be the most effective.

The parameters were applied to the proposed composite models and evaluated for accuracy with respect to the baseline model, as shown in Table 2. The composite models have a lower MSE per frame with the unconditioned model performing slightly better.

**Optimization and Initialization** The cost function of eq. (6) was optimized with RMSProp [22]. RMSProp uses a running average over the root mean square of recent gradients to normalize the gradients and divide the learning rate. A learning rate of $10^{-4}$ and decay rate of 0.9 were used. Adagrad [23] and Adam [24] were both considered and tested as well, but RMSProp was empirically chosen as the most effective. The learning rate was set to $10^{-4}$. We used a mini-batch of five video sequences and trained the models for up to 25,000 iterations. Early stopping was performed based on the validation loss if necessary. The weights were initialized using the Xavier algorithm [25]. It automatically scales the initialization based on the number of input and output neurons to prevent the weights from starting out too small or large, and vanish or explode in magnitude. The input-to-hidden and hidden-to-hidden convolutional filters in the Conv-LSTM units all use the same filter size.

Table 1: Baseline model parameter selection

| Parameter Selection: Input Length, Output Length, Filter Size, Output Nonlinearity | MSE Per Frame |
|---|---|
| 5 , 5 , (5x5) , sigmoid | 20.51 |
| 10 , 5 , (5x5) , sigmoid | 17.47 |
| 5 , 10 , (5x5) , sigmoid | 23.03 |
| 10 , 10 , (5x5) , sigmoid | 24.86 |
| 5 , 5 , (3x3) , sigmoid | 30.18 |
| 5 , 5 , (5x5) , ReLU | 24.83 |

Table 2. Model Accuracy Comparisons

| Model | MSE per Frame |
|---|---|
| Baseline | 20.51 |
| Composite Encoder-Decoder | 14.83 |
| Conditioned Composite Encoder-Decoder | 16.07 |

**Evaluation Parameters** A temporal window of fifty frames before and after distinct local minima is used to propose anomalous regions, as most anomalous activities are at least one hundred frames long. The proposed regions of local minima within fifty frames of one another are joined to obtain the final abnormal frame regions. These minima are then considered to be a part of same abnormal event. We consider a detected abnormal region as a correct detection if it has at least fifty percent overlap with the ground truth. A parameter-sweep at intervals of .05 is performed to determine the threshold parameter for the Persistence1D algorithm [21].

## 4.3. Predicting the Future of a Video Sequence

The learned models successfully reconstruct the past and predict the future. An example of reconstruction and prediction of both a regular and anomalous video sequence is depicted in Figure 5. It can be seen that the reconstruction of pedestrians walking, a normal event, within the anomalous sequence is portrayed correctly by the model, while the small vehicle (anomaly) loses detail with each timestep. The past reconstruction of the vehicle is noisy but more accurate because all of the information is already within the encoding. The future prediction of the vehicle's motion deteriorates noticeably at each timestep. In fact, because the model only learned to extrapolate the motion of pedestrians, the shape of the vehicle begins to look increasingly similar to a pedestrian over time. The visualization of output video sequences makes it clear that the model will only be able to accurately reconstruct and predict 'normal' events that were learned during training.

Table 3: Comparing anomaly detection performance of the proposed models on UCSD Pedestrian 1 dataset

|  | TP/FP | Precision | Recall |
|---|---|---|---|
| Ground Truth | 40/0 | 1.00 | 1.00 |
| Composite Encoder-Decoder (224x224) | 35/8 | .81 | .88 |
| Conditioned Composite Encoder-Decoder (224x224) | 35/7 | .83 | .88 |
| State of the Art Hasan et al. [15] | 38/6 | .864 | .95 |
| Composite Encoder-Decoder (64x64) | 40/7 | .851 | 1.00 |
| Conditioned Composite Encoder-Decoder (64x64) | 39/7 | .848 | .975 |

Table 4: Average MSE per frame when evaluated on 64x64 pixel images using the specified parameters

|  | 3x3 filter | 5x5 filter |
|---|---|---|
| Composite Encoder-Decoder | 0.5697 | 0.3469 |
| Conditioned Composite Encoder-Decoder | 0.6872 | 0.4136 |

## 4.4. Anomalous Event Detection

The proposed models were coded using Lasagne [26]. Initial training and evaluation were done on the UCSD Pedestrian 1 dataset. The best model was improved upon and further evaluated on the remaining datasets.

**UCSD Pedestrian 1**  This dataset contains a variety of abnormal events that can be categorized into two main categories, the movement of non-pedestrian entities and anomalous pedestrian motions. Common anomalies include bikers, skaters, and people walking perpendicular to the walkway, or off the walkway. The crowd densities range from sparse to very crowded, providing a wide range of regular actions. The dataset is split into training and testing data. No anomalies occur within the training data, while at least one anomaly occurs per testing clip.

The anomaly detection results for the proposed models, with 224x224 input size are shown in Table 3. Both the conditioned and unconditioned versions of the composite encoder-decoder model exhibit similar detection rates.

**Improving Performance**  We explored ways to improve the performance of the composite Encoder-Decoder model and found that compressing the size of the input is useful. An improved model with input image size of 64x64 pixels is considered. The downsized 64x64 image is split into 16 non-overlapping patches, instead of 64 patches for the case of 224x224 image. We trained the model with the baseline 5x5 filter as well as a smaller 3x3 filter, since the amount of motion is reduced in the downsized image. Based on the results of Table 4, we selected the 5x5 filter because it provided the smallest reconstruction error.

The improved Encoder-Decoder model is shown to outperform the state of the art method of [15] (see Table 3). The unconditioned model has one more true positive detection than the conditioned version, increasing its recall rate to 100%. It should be noted that while the proposed model yields seven false positives, a closer examination shows that the ground truth annotation of UCSD Pedestrian 1 does not account for several instances of pedestrians walking off the walkway. Furthermore, when detecting anomalies in surveillance environments the cost of missing anomalies is much higher than the cost of false positives. Thus, it is more desirable to detect more anomalies, even if false positives increase slightly.

As the Unconditioned Composite Conv-LSTM Encoder-Decoder model exhibits the best performance, we further evaluate it on the rest of the datasets for a comprehensive quantitative evaluation shown in Table 5. To make the results easier to visualize, the regularity score evaluation is graphed, as shown in Figure 6. Distinct local minima and maxima are represented by a blue dot and a red dot respectively. The anomalous ground truth regions are highlighted in red, and the proposed anomalous regions are highlighted in green. It should be noted that the last nine frames of each video sequence are not evaluated since the model requires a minimum of ten frames, five as the input and five as the target, to return a reconstruction score.

**UCSD Pedestrian 2**  This dataset is similar in content to UCSD Pedestrian 1. However, it features a different walkway, and contains fewer anomalies. There is only one abnormal event per testing clip, with the anomaly spanning the majority of the video segment. The proposed model is able to learn the regular motion of pedestrians walking and differentiate it from any abnormalities. There is a 100% recall rate with only one false positive on this dataset.

Table 5: Comparing abnormal event detection performance across multiple datasets. Ours is the Composite Conv-LSTM Encoder-Decoder (64x64) model. [*] Improved ground truth on Subway Exit dataset. [#] Uses older dataset.

| Dataset | | | Anomaly Detection | | | | | | | |
|---|---|---|---|---|---|---|---|---|---|---|
| Name | # Frames | # Anom alies | True Positive/False Alarm | | | | Precision / Recall | | | |
| | | | Ours | [15] | [14] | [9] | Ours | Hasan [15] | Lu [14] | Zhao [9] |
| UCSD Ped 1 | 7,200 | 40 | 40/7 | 38/6 | N/A | N/A | .851 / 1.00 | .864 / .95 | N/A | N/A |
| UCSD Ped 2 | 2,010 | 12 | 12/1 | 12/1 | N/A | N/A | .923 / 1.00 | .923 / 1.00 | N/A | N/A |
| Subway Enter | 121,749 | 66 | 62/14 | 61/15 | 57/4 | 60/5 | .816 / .939 | .803 / .92 | .934 / .864 | .923 / .909 |
| Subway Exit | 64,901 | 19 | 19/37 | 17/5 | 19/2 | 19/2 | .339 / 1.00 | .773 / .895 | .904 / 1.00 | .904 / 1.00 |
| | | 30[*] | 29/15 | N/A | N/A | N/A | .659 / .967 | N/A | N/A | N/A |
| Avenue | 15,324 | 47 | 40/2 | 45/4 | 12/1[#] | N/A | .952 / .923 | .918 / .957 | .923 / N/A | N/A |

**Subway Entrance** The subway entrance surveillance video contains a variety of anomalous activities that include commuters walking in the wrong direction, avoiding payment and loitering. The first 15-minutes of this one-hour video are used for training. The last few seconds of the video are clipped off since the camera moves and it no longer pertains to the entrance. Our approach outperforms other methods by detecting 62 of 66 anomalies with a 93.9% recall.

**Subway Exit** The subway exit surveillance video contains a variety of anomalies similar to those found in the Subway Entrance video. The first 15 minutes are used for training as well. Unlike the former video, the training data is not a good representation of usual events, with the majority of all action taking place at two short intervals and only one example of people exiting the station. Furthermore, there are several variations of normal events that occur in the testing data that are not present during training. This results in a skewed model that cannot accurately model the regularity of a video. Another issue is the fact that the ground truth defines only 19 anomalous events when closer inspection reveals 30.

Our evaluation included both the ground truth as provided in [15] and an updated one that better labels all anomalies. As seen in Table 5, the model is able to detect every anomalous occurrence in both ground truth annotations. Some of the original false alarms are actually regions where an anomalous event takes place, and performance improves significantly when the model is evaluated using the updated ground truth annotation,. The total number of detected anomalies decreases, as many of the previous false alarms fall under the same anomalous event that is only counted once.

**Avenue** The Avenue dataset is split into training and testing data, with each video around 2 minutes long. The testing video clips contain a wide variety of anomalous events including but not limited to running, thrown objects, and walking in the wrong direction. The training video clips contain mostly 'normal' activity, but do include a few irregular events. The proposed model is able to precisely differentiate between normal and anomalous activity in this dataset. However, it fails to detect several anomalous events of jogging that occur in the background where most of the 'normal' action takes place and the pedestrians appear smaller. Since the deviation in regularity caused by a smaller object in the background is less significant than larger or more disruptive anomalous events, the evaluation algorithm is unable to differentiate the action of jogging from walking pedestrians.

## 5. Conclusions

We present a composite Conv-LSTM network architecture that is able to model video sequences, perform reconstruction, and predict future frames. We apply the predictive capabilities of our network to determine anomalous events and locate points of interest in video sequences. A comparison of predictions between normal and anomalous events shows that our Conv-LSTM networks accurately model learned (familiar) movements but do not adapt to new (unusual) movements. This makes such networks effective for recognizing abnormalities when the training data is loosely supervised to contain mostly 'normal' events. A quantitative analysis shows that our model performs competitively with state-of-the-art anomaly detection methods on various datasets.

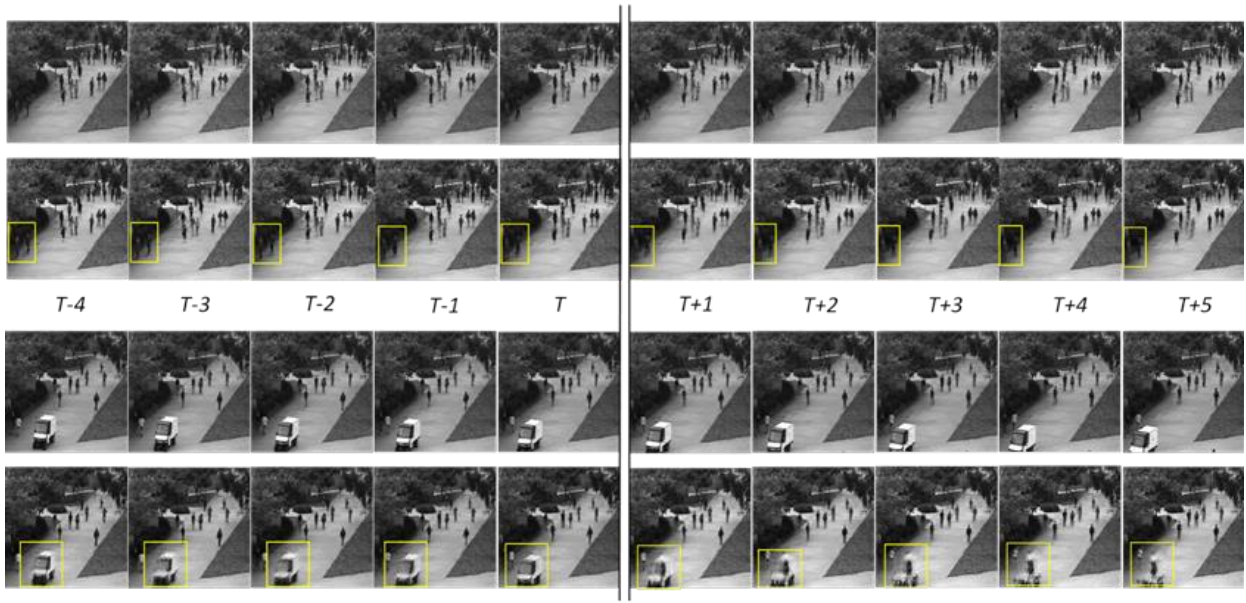

Figure 5. Reconstruction and future prediction obtained from the Composite Conv-LSTM Encoder-Decoder Model on two UCSD Pedestrian 1 sequences. The first and third rows are the ground truth video sequences, while the second and fourth rows are the model reconstruction and prediction. Each column denotes a timestep, with the first five corresponding to the past input, and the last five corresponding to the future predictions. Both the reconstruction and prediction begin at *T*. Only the video sequence shown in the third and fourth rows contain an anomaly. Regions of interest are highlighted by a yellow outline. There are two pedestrians and a small vehicle within the outlines of the second and fourth row, respectively. Additional examples are presented in the supplementary material.

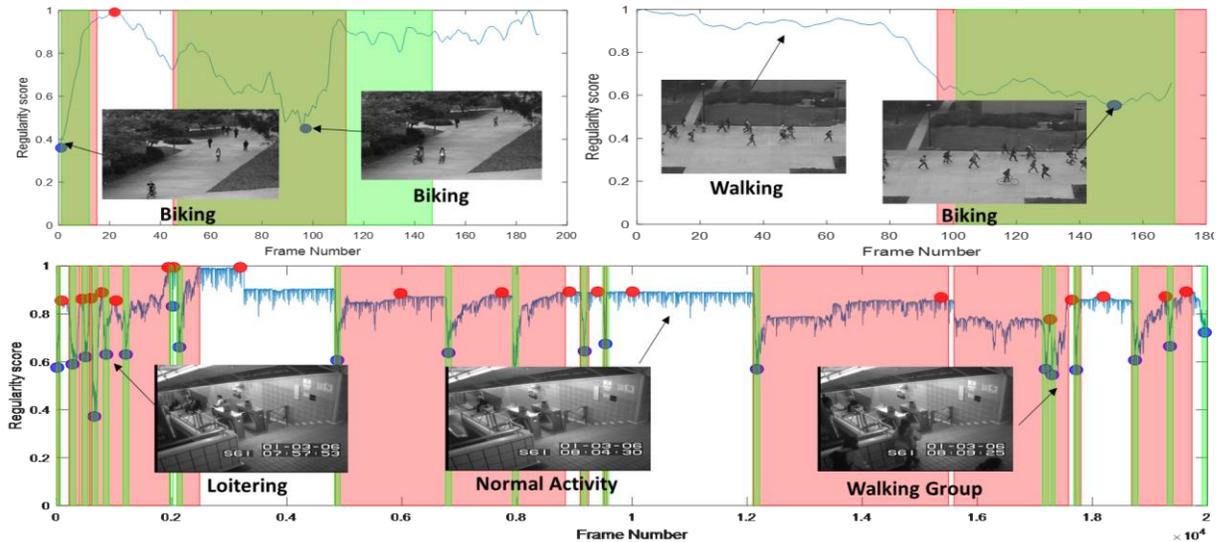

Figure 6. Regularity score of each frame of three video sequences. (Top-Left) UCSD Pedestrian 1 Testing Clip #29, (Top-Right) UCSD Pedestrian 2 Testing Clip #2, (Bottom) Subway Exit of Frames 20,000 − 40,000. The Frame Number indicates the starting frame of the input video sequence used during testing to obtain a regularity score. Additional examples are shown in the supplementary material.

# Anomaly Detection in Video Using Predictive Convolutional Long Short-Term Memory Networks

Jefferson Ryan Medel and Andreas Savakis

Rochester Institute of Technology

## Table of Contents

# 1. Video Input Reconstruction and Future Prediction

Our composite Conv-LSTM Encoder-Decoder network is described in the paper submission and shown in Figure 1 below. A regularity score used to perform anomaly evaluation is computed from the the output reconstruction and prediction errors. To better understand what properties of a video sequence produce a high or low regularity score, we provide a visualization of a composite model. Since the 64x64 images are harder to visually evaluate due to their low resolution, we present a visualization of the outputs for our (224x224) composite Conv-LSTM encoder-decoder model. The output for anomalous video sequences shows that the trained model is unable to correctly reconstruct or predict anomalous sequences and in fact distorts them to better resemble normal events. The input reconstructions are always better than the future predictions because the encoding already contains all of the information about the input.

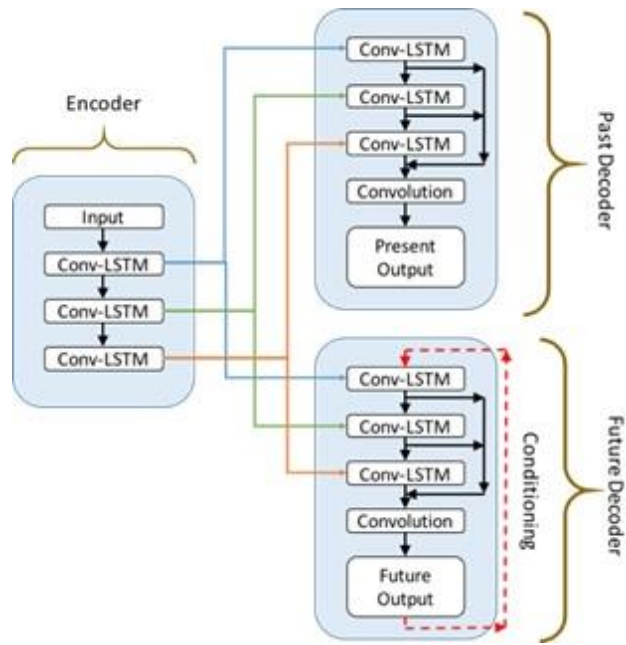

Figure 1. High-level view of the Conditioned Composite Conv-LSTM Encoder-Decoder.

For Figures 2 to 11, each column denotes a time-step; the first row and third row show the ground truth; the second row is the reconstruction of the input and the fourth row is the future prediction of the video. Regions of interest are highlighted by a yellow bounding box.

## *1.1.   Non-Anomalous Video Sequences*

### 1.1.1   UCSD Pedestrian 1

Figure 2 depicts pedestrians normally walking down a walkway. The input reconstruction for is nearly identical to the target ground truth. Even though the future prediction of the output degrades in quality over time, the silhouette of the highlighted pedestrian is distinct enough to make out each body part.

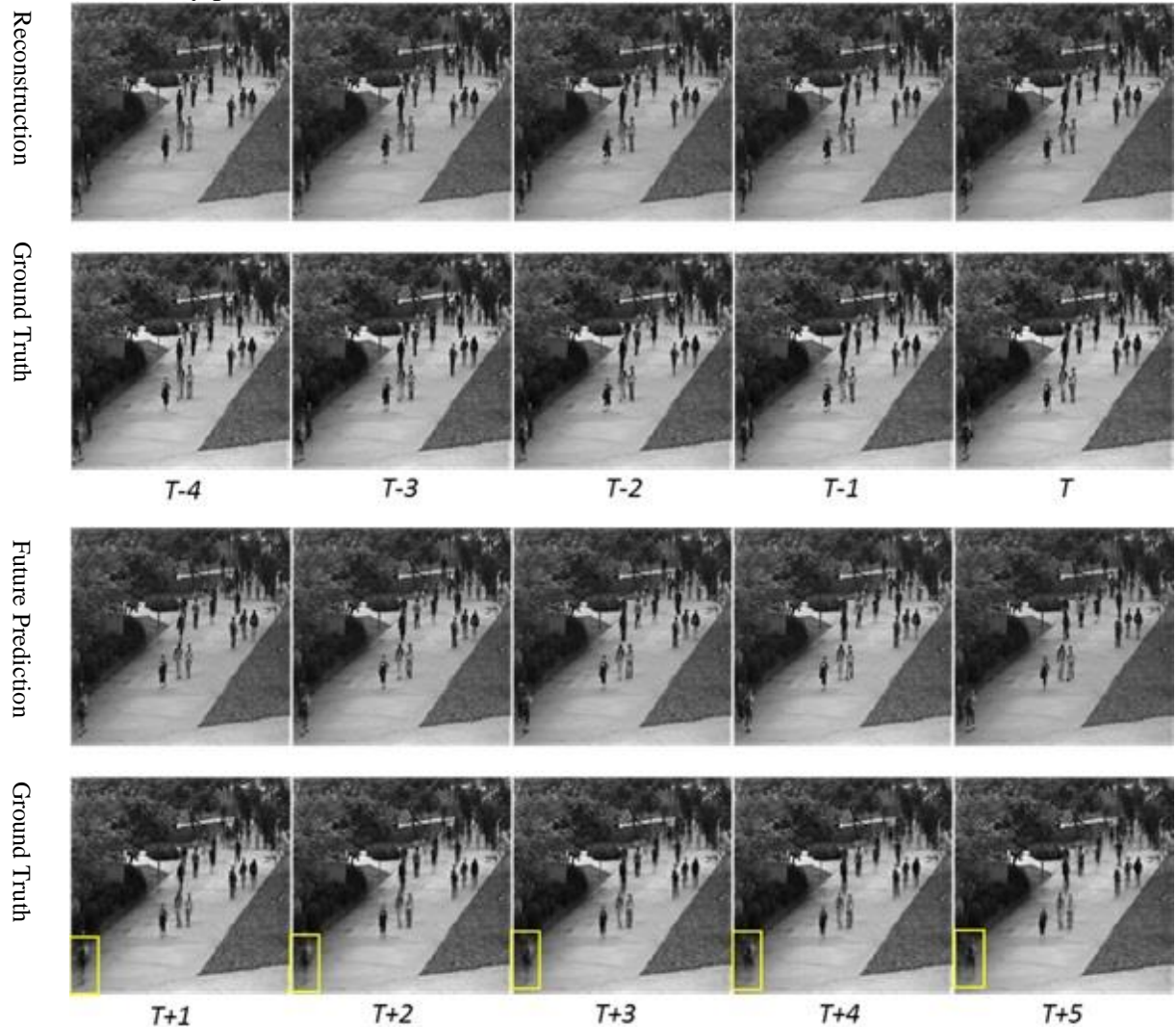

Figure 2. A visualization of the output from the (224x224) Composite Conv-LSTM Encoder-Decoder Model on a *non-anomalous* sequence from test clip 1 of the UCSD Pedestrian 1 dataset.

### 1.1.2 UCSD Pedestrian 2

Figure 3 depicts pedestrians walking in parallel down a walkway. Both the input reconstruction and future prediction seen within Figure 3 are very similar to the ground truth. The future prediction is worse than the reconstruction, as it contains increasing amounts of blur with every subsequent timestep. However, every pedestrian is distinct enough from the others despite being close together.

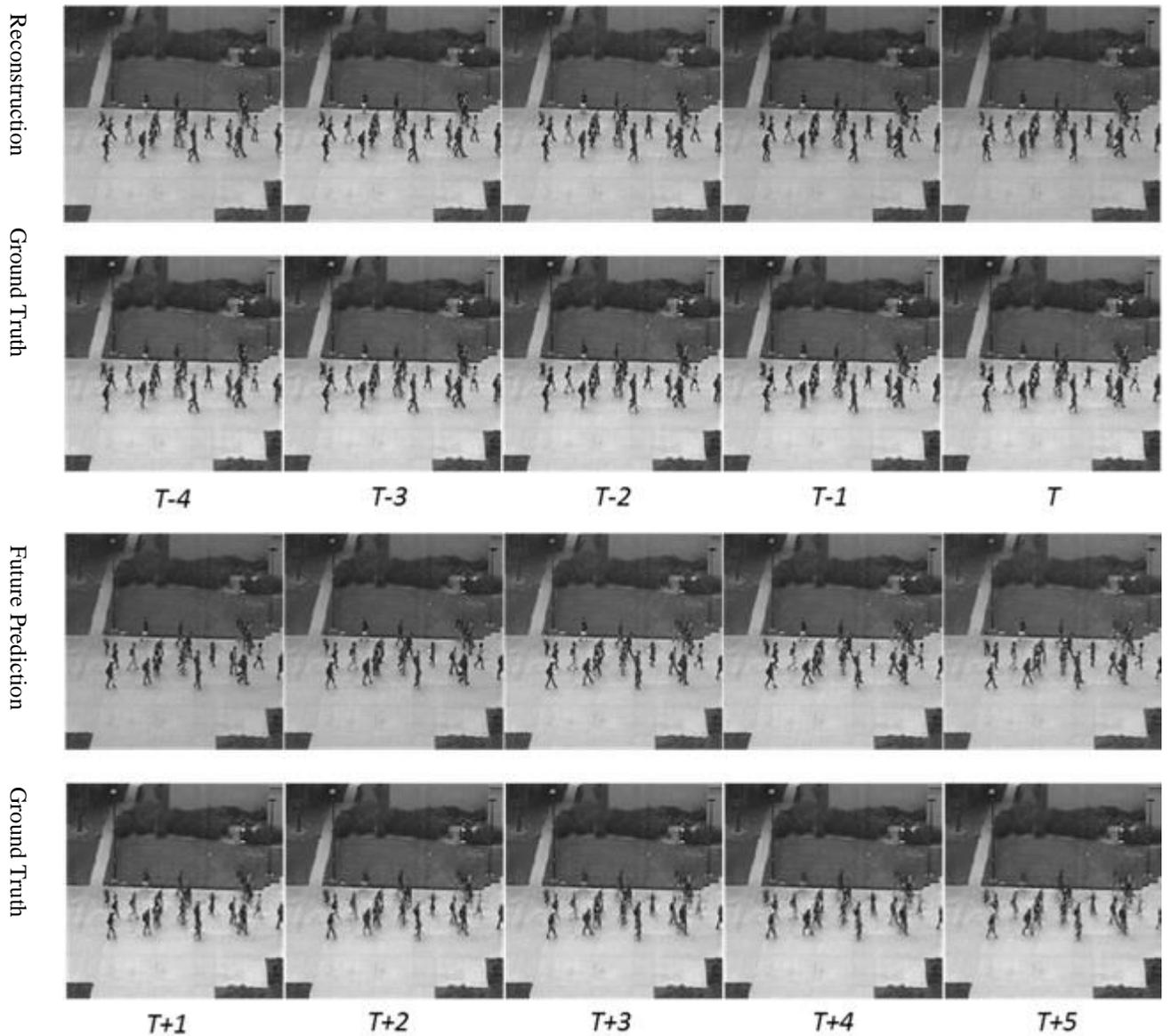

Figure 3. A visualization of the output from the (224x224) Composite Conv-LSTM Encoder-Decoder Model on a *non-anomalous* sequence from test clip 1 of the UCSD Pedestrian 2 dataset.

### 1.1.3 Subway Entrance

Figure 4 shows people at a train station waiting to board the train. As with Figures 2 and 3, the input reconstruction in Figure 4 is very accurate. The station itself is accurate in the future prediction, while the person highlighted begins to fade slightly.

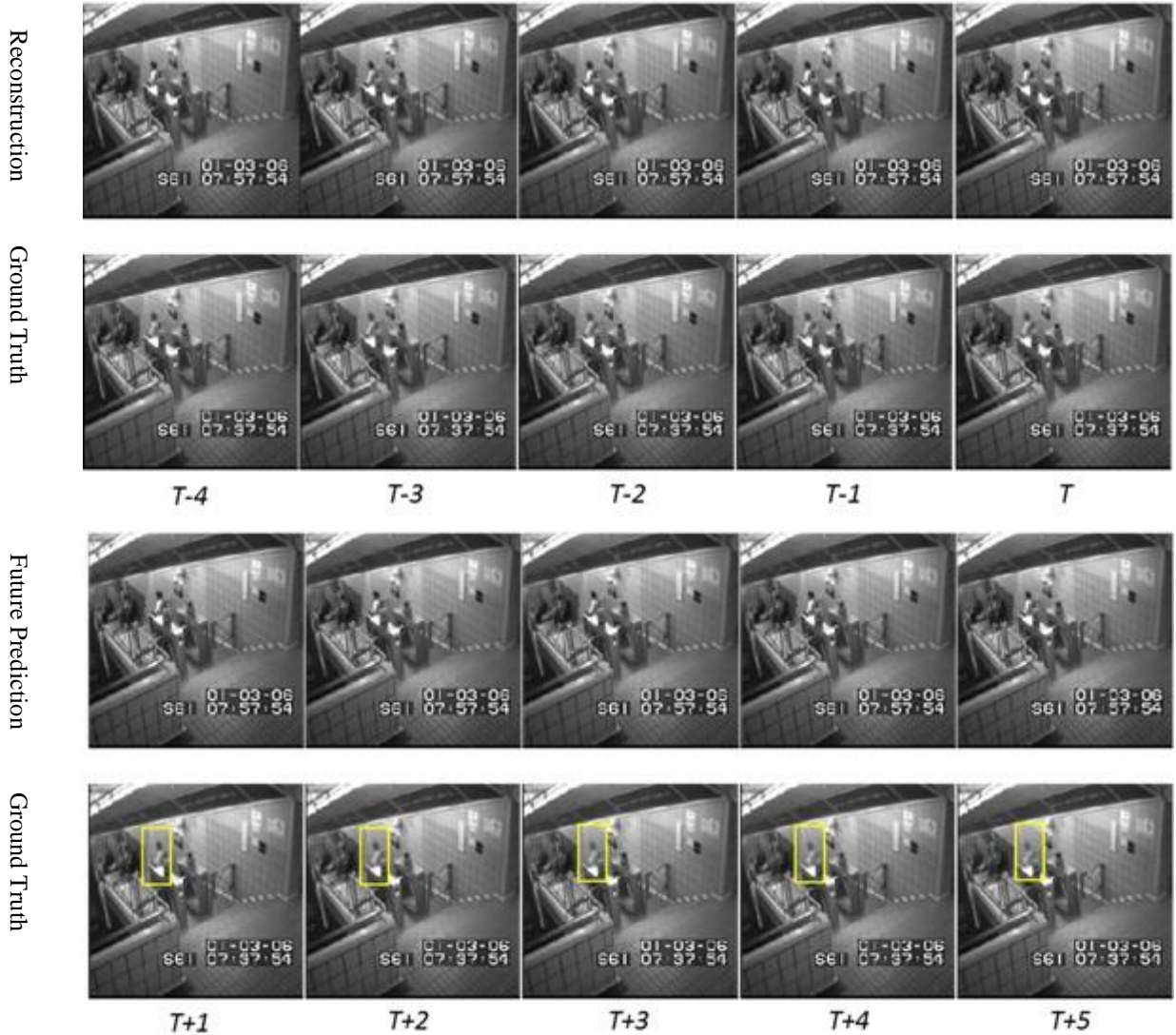

Figure 4. A visualization of the output from the (224x224) Composite Conv-LSTM Encoder-Decoder Model on a *non-anomalous* sequence from the Subway Entrance Video.

## 1.1.4 Subway Exit

Figure 5 shows two people at a train station that are about to leave. The input reconstruction shows slightly less resolution than the ones previously seen. It is, however, still very accurate. As with Figure 4, the people within the future prediction begin to blur slightly with every timestep.

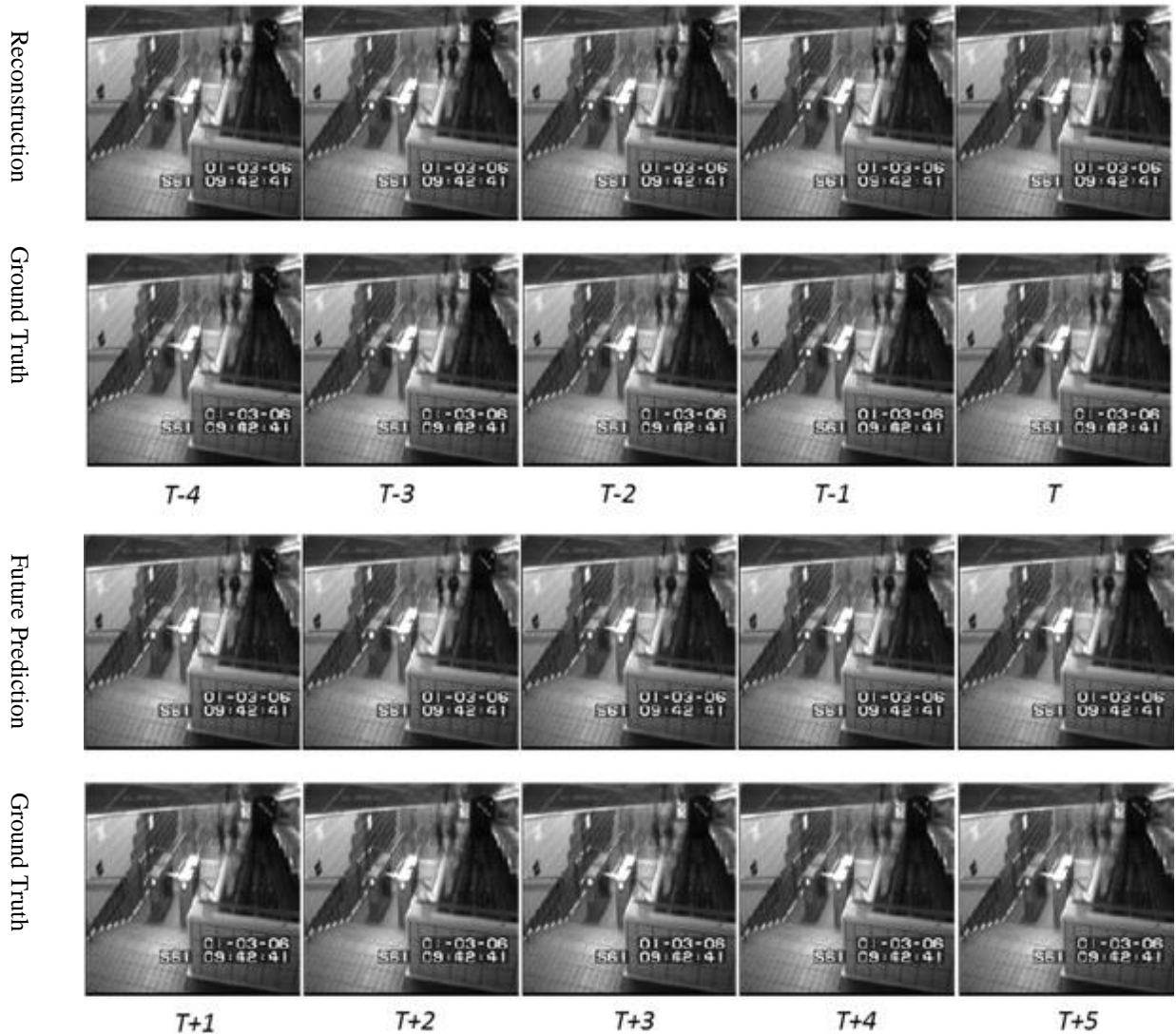

Figure 5. A visualization of the output from the (224x224) Composite Conv-LSTM Encoder-Decoder Model on a *non-anomalous* sequence from the Subway Exit Video.

## 1.1.5 Avenue

Figure 6 depicts a crowd of pedestrians walking in parallel below the archway. Like previous figures in this section, the input reconstruction is very accurate. The pedestrians in the future reconstruction, are slightly more blurred than the ones seen in Figures 2 to 5.

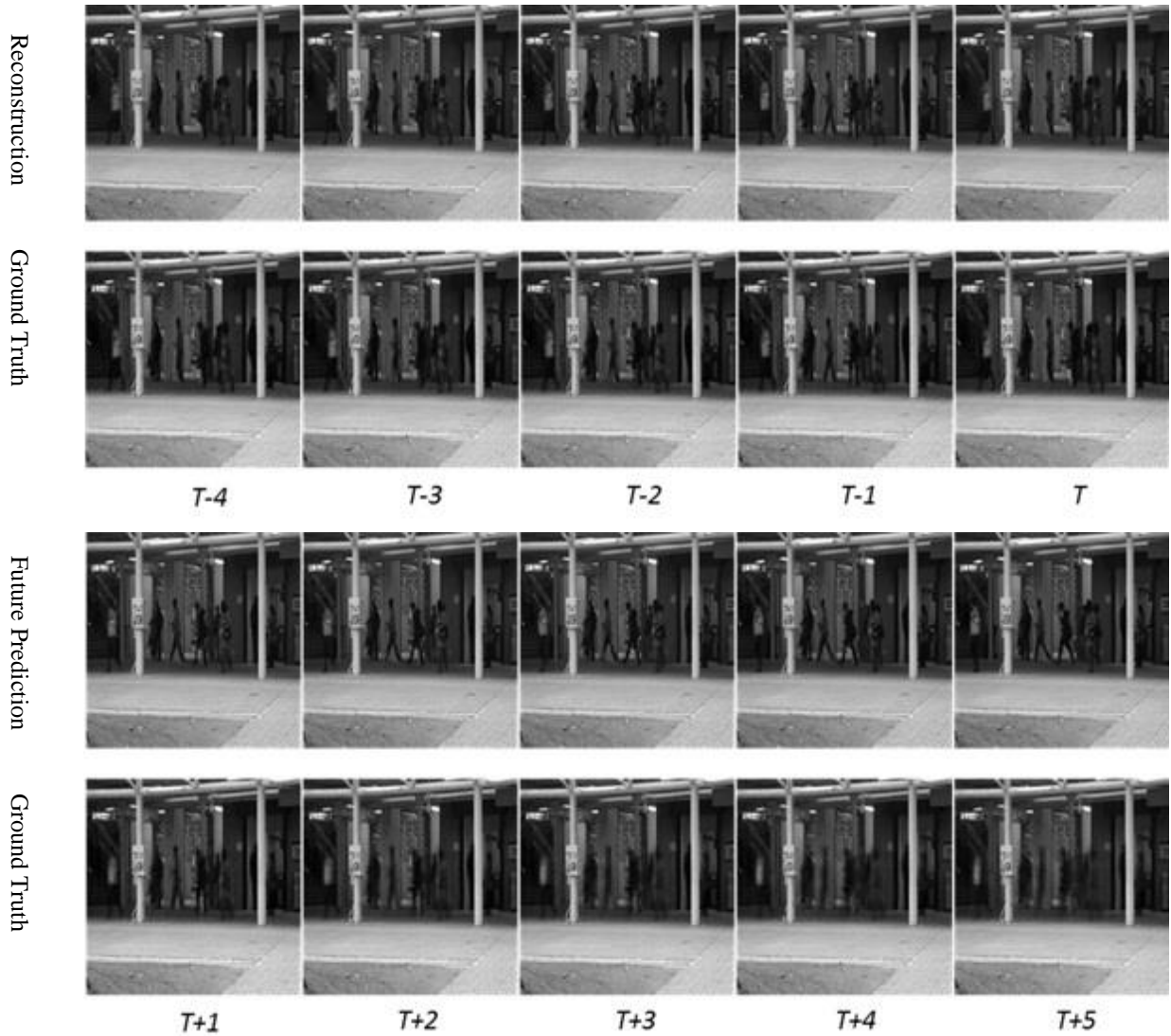

Figure 6. A visualization of the output from the (224x224) Composite Conv-LSTM Encoder-Decoder Model on a *non-anomalous* sequence from test clip 4 of the Avenue dataset

## *1.2. Anomalous Video Sequences*

### 1.2.1 UCSD Pedestrian 1

A person riding a bicycle on the walkway is seen in Figure 7. The input reconstruction of the biker contains a slight distortion not seen with the other pedestrians. Furthermore, the bike seems to merge with the person in the future reconstruction, showing the model's rejection of anomalous events.

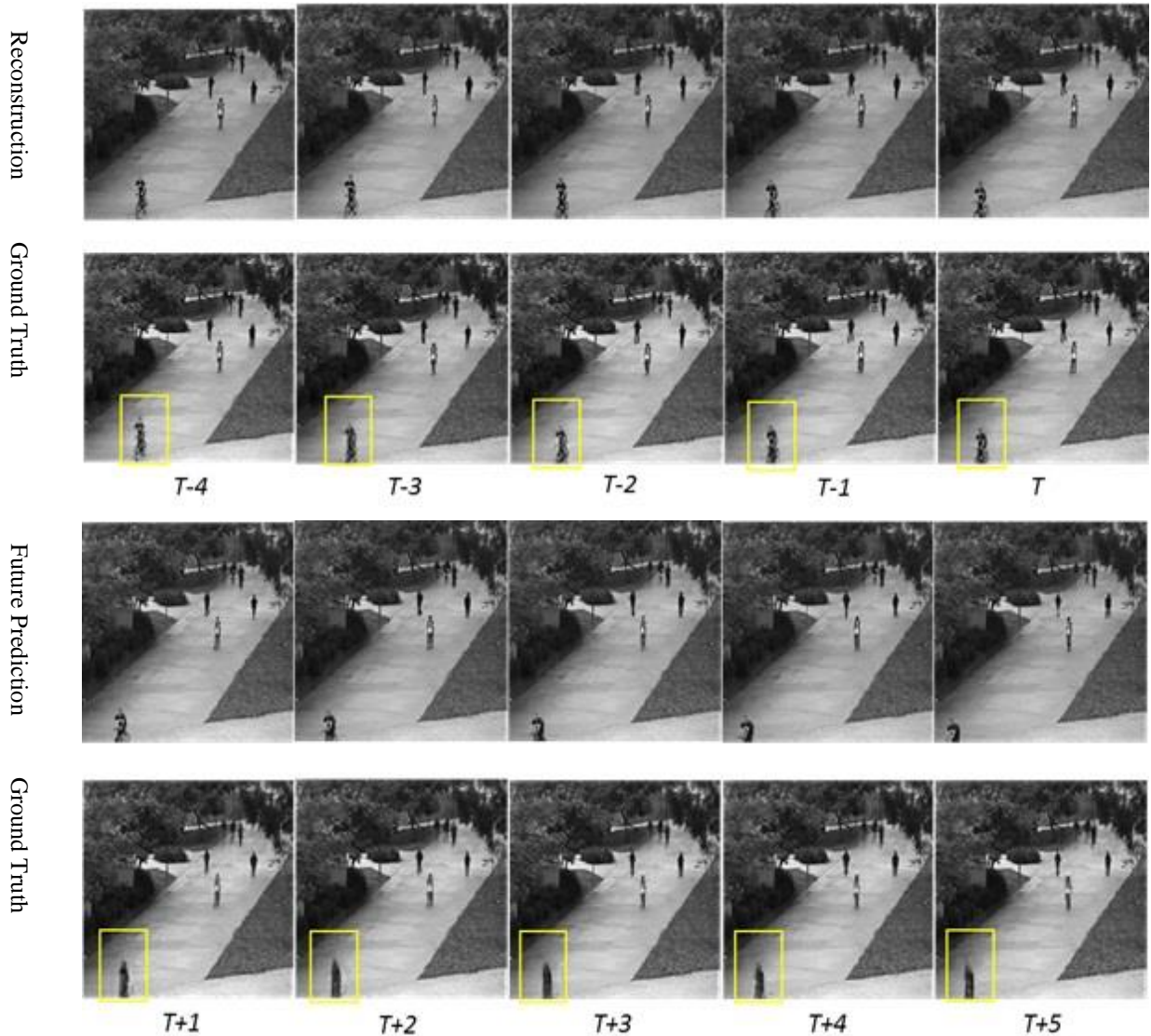

Figure 7. A visualization of the output from the (224x224) Composite Conv-LSTM Encoder-Decoder Model on an *anomalous* sequence from test clip 29 of the UCSD Pedestrian 1 dataset.

### 1.2.2 UCSD Pedestrian 2

Figure 8 contains a person riding a bicycle down the walkway. The shape of the bicycle is not completely reconstructed in each frame of the input reconstruction, and merges with the person within the future prediction. By T+4, the bicycle is gone altogether.

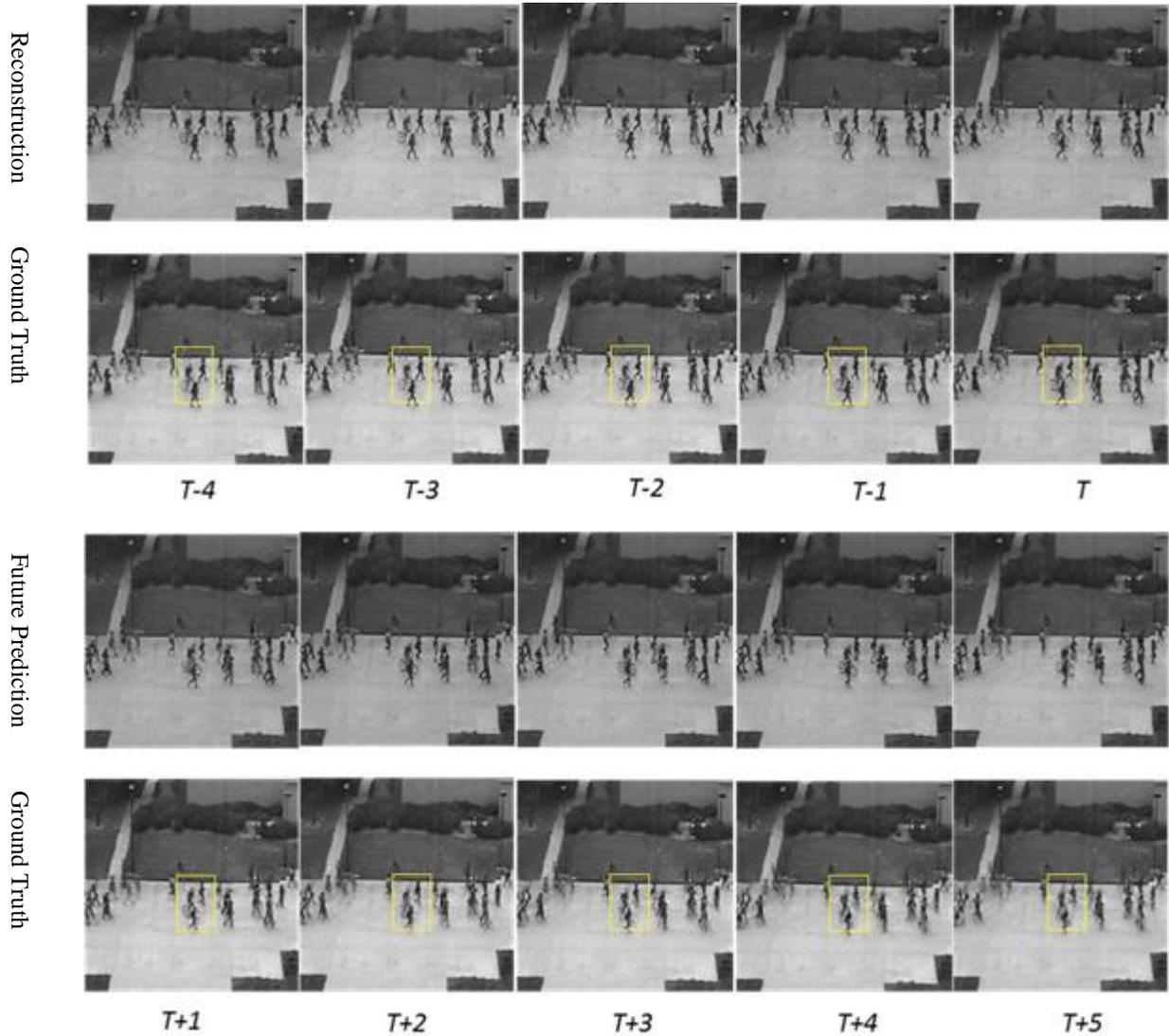

Figure 8. A visualization of the output from the (224x224) Composite Conv-LSTM Encoder-Decoder Model on an *anomalous* sequence from test clip 29 of the UCSD Pedestrian 1 dataset.

## 1.2.3 Subway Entrance

Figure 9 shows three people walking across the entrance. It can be seen that the silhouettes of the three people walking across visibly distort while the ones of the people waiting on the platform do not.

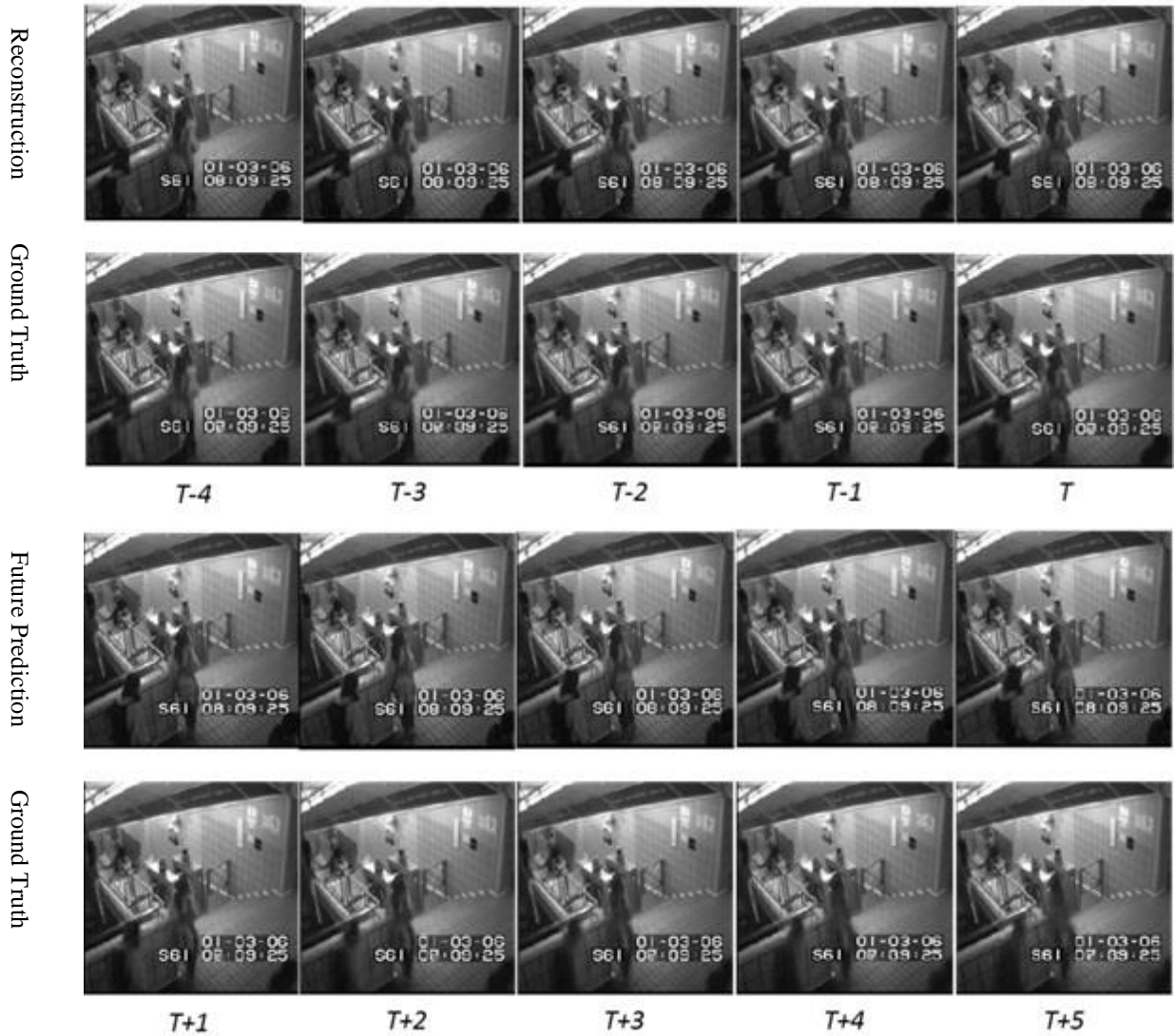

Figure 9. A visualization of the output from the (224x224) Composite Conv-LSTM Encoder-Decoder Model on an *anomalous* sequence from Subway Entrance video.

## 1.2.4  Subway Exit

Figure 10 shows a crowd of people leaving the platform, with a single person, highlighted by the yellow bounding box, trying to enter it. While the people within the crowd blur for both the input reconstruction and future prediction, the person trying to enter does not move within the reconstruction and disappears within the prediction.

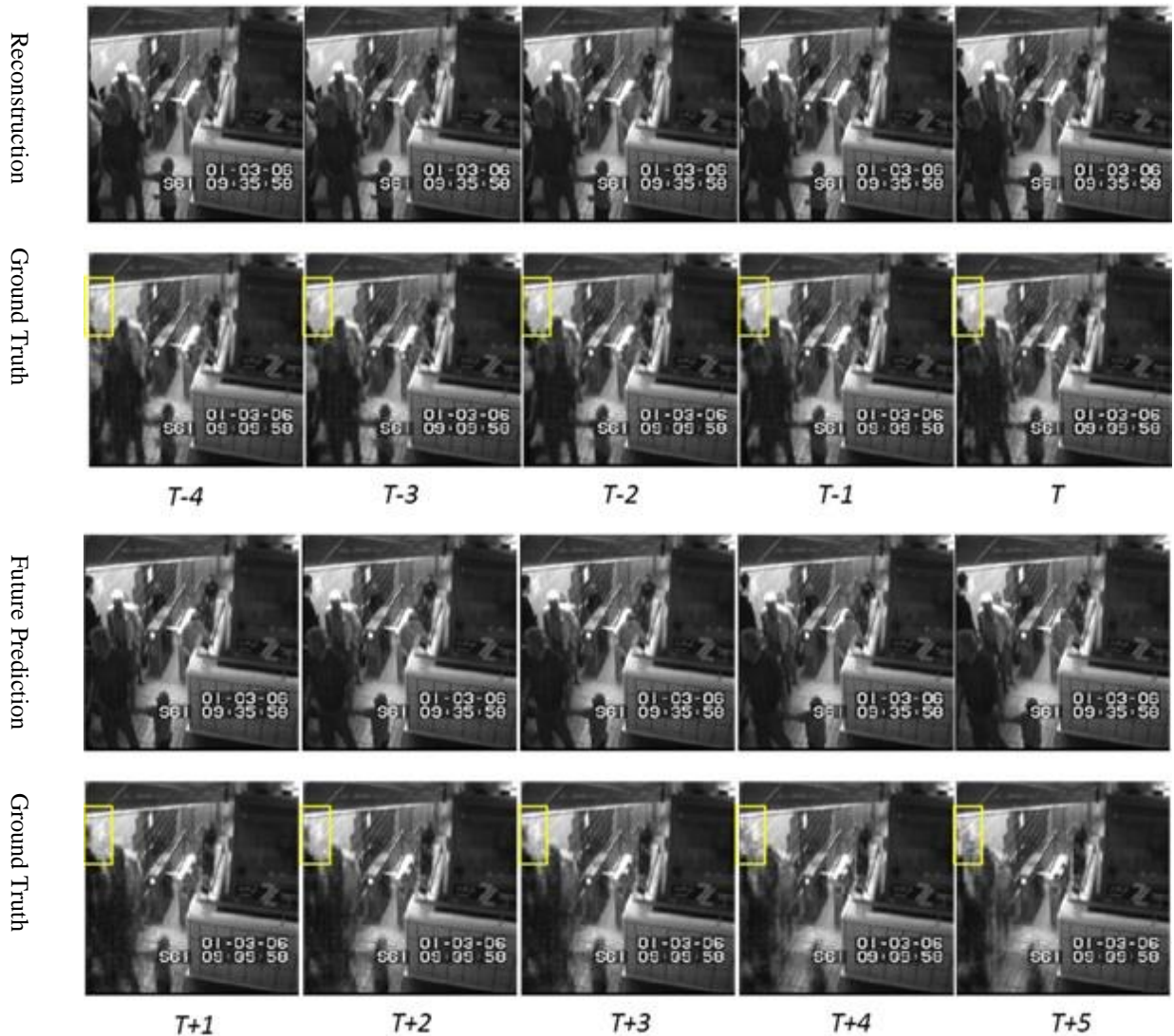

Figure 10. A visualization of the output from the (224x224) Composite Conv-LSTM Encoder-Decoder Model on an *anomalous* sequence from Subway Exit video.

## 1.2.5 Avenue

Figure 11 contains a person walking in the wrong direction, perpendicular to the walkway, and very close to the camera. The input reconstruction of the person is noticeably distorted, while the future prediction blurs out any of her distinct features. It can be seen that the pedestrians in the background are correctly modeled.

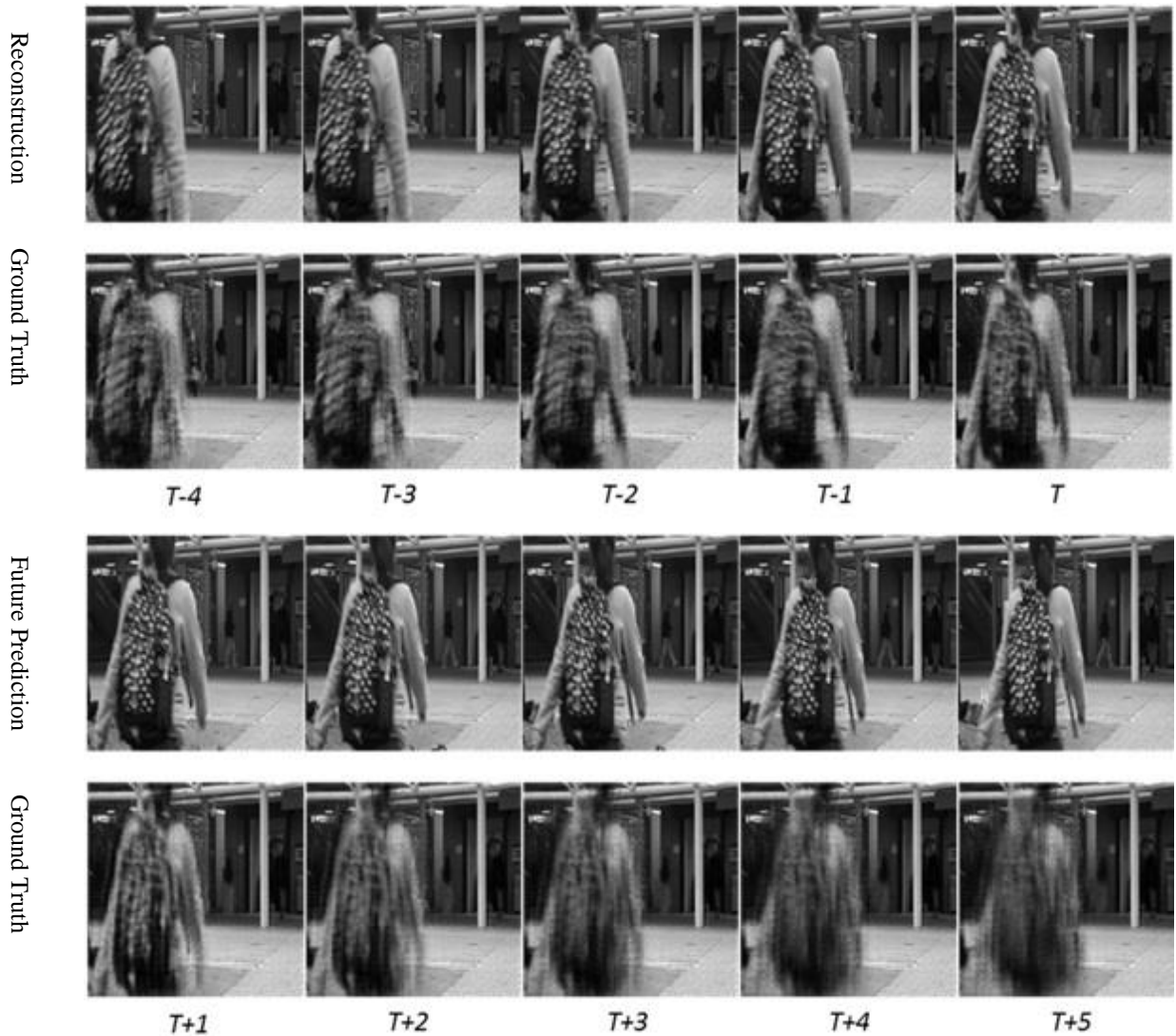

Figure 11. A visualization of the output from the (224x224) Composite Conv-LSTM Encoder-Decoder Model on an *anomalous* sequence from test clip 1 of the Avenue dataset.

## 2.      Anomaly Detection Evaluation

A visualization of the anomaly evaluation using the regularity scores of the input video sequences is depicted in Figures 12 to 16 for each dataset. The graphed line is the regularity score per input video sequence, starting at the value denoted by the x-axis. The red and green shaded regions denote the anomalous ground truth and prediction, respectively. The red and blue dots denote the distinct local maxima and minima, respectively. It can be seen that dips in the regularity score correspond well with the target predictions. This shows that the learned models are able to capture the normality of training videos well enough to understand which events are not consistent with typical events and can be considered anomalies. The anomalies create a lower regularity score the closer it is to the camera. This is due to the fact that events that occur closer to the camera will be larger, and take up more area within the video frame. This in turn increases the reconstruction and prediction errors of the video sequence.

Each graph is accompanied by one or more video frames showing what exactly causes the given regularity score. Yellow bounding boxes are used to highlight objects that may be difficult to find.

## 2.1. UCSD Pedestrian 1

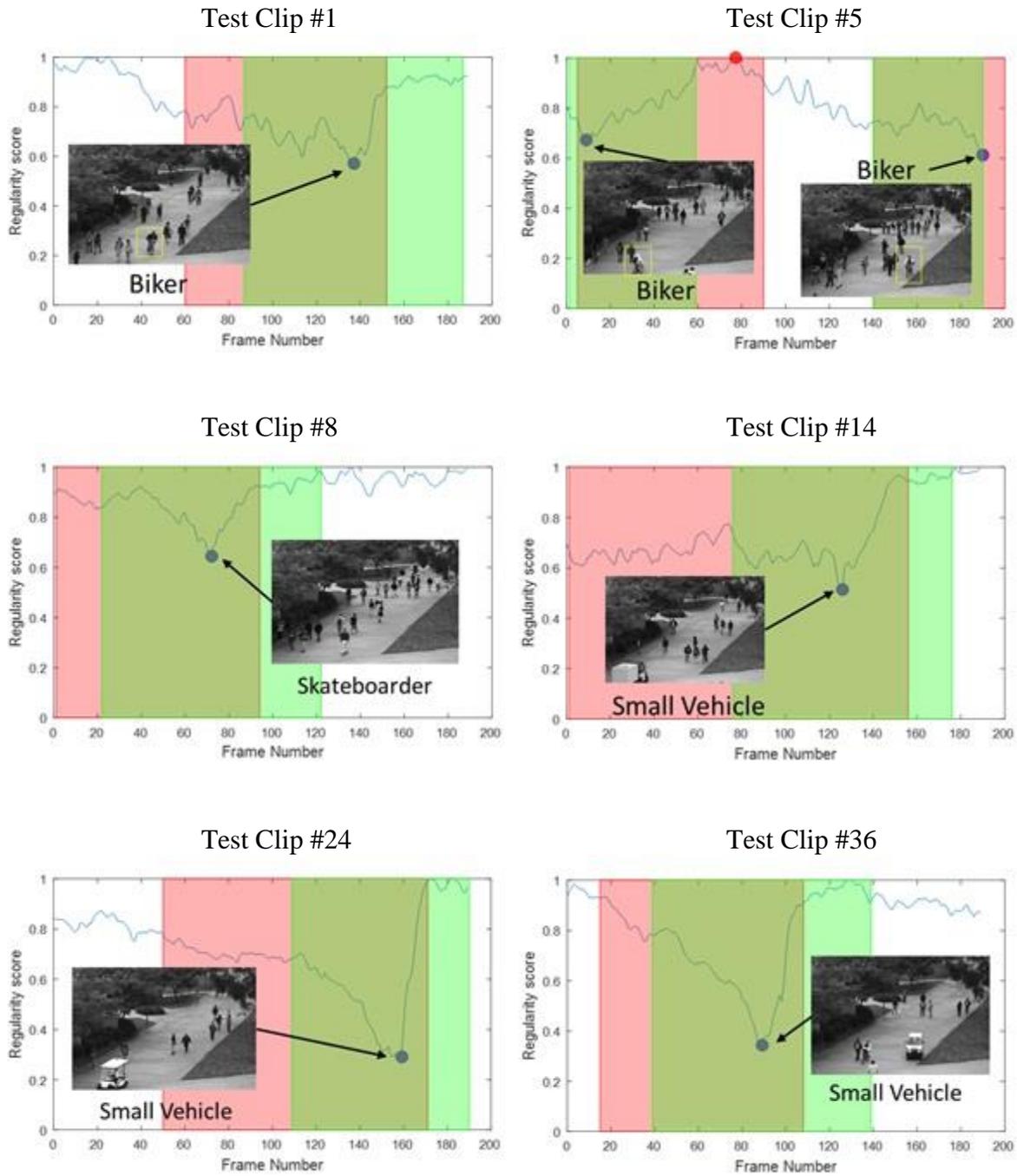

Figure 2. Anomaly evaluation graphs of test clips from the UCSD Pedestrian 1 dataset.

## 2.2. UCSD Pedestrian 2

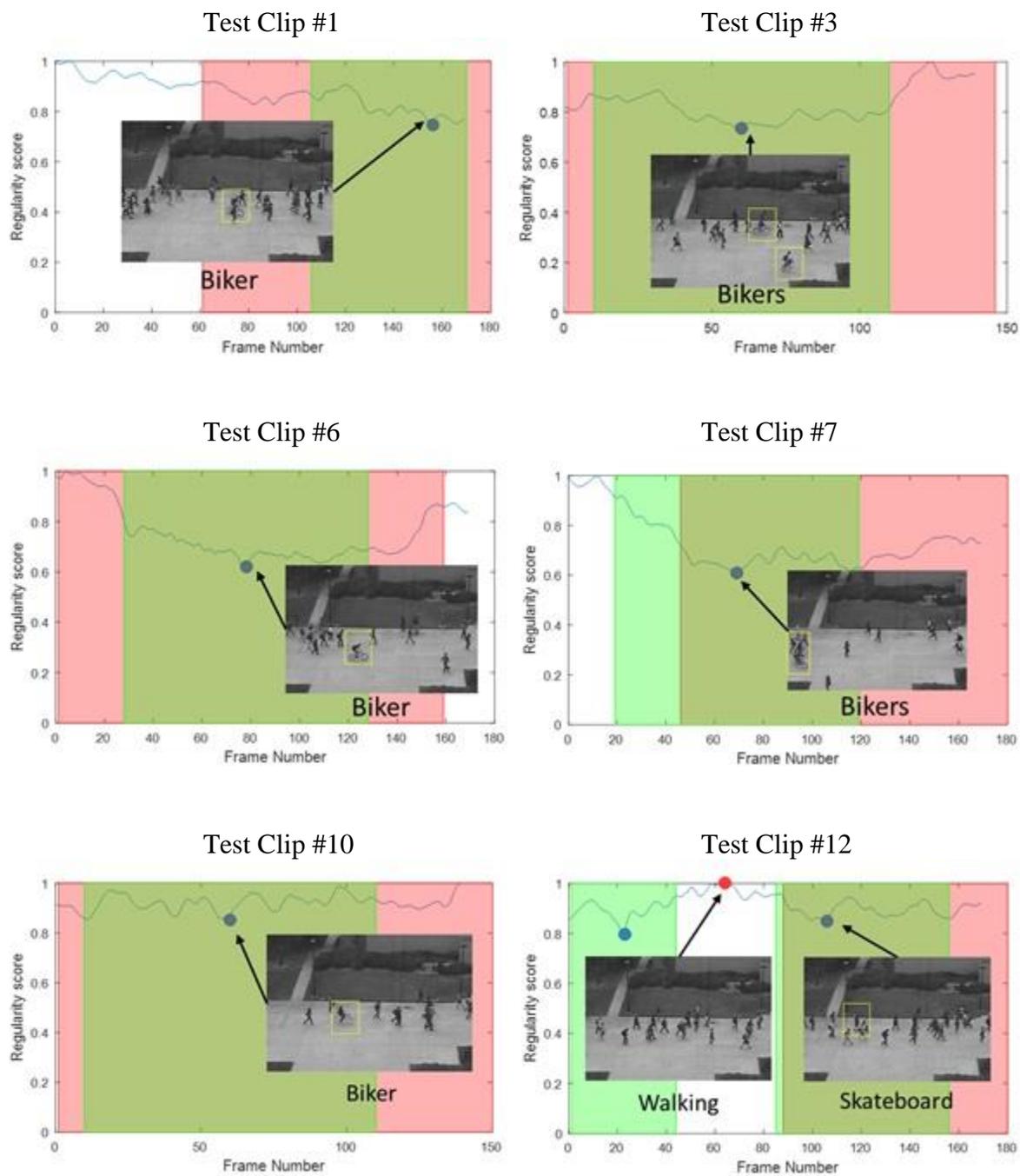

Figure 13. Anomaly evaluation graphs of test clips from the UCSD Pedestrian 1 dataset.

## 2.3. Subway Entrance

The anomalous events within this video are generally longer than the window used to predict anomalous segments. However, we continue to utilize a 100 frame window, because there are several events that are very short. Furthermore, it is evident that the predicted anomalous segments still mostly occur within the target anomalous regions, allowing the anomalies to be found.

Frames 100,000-120,000

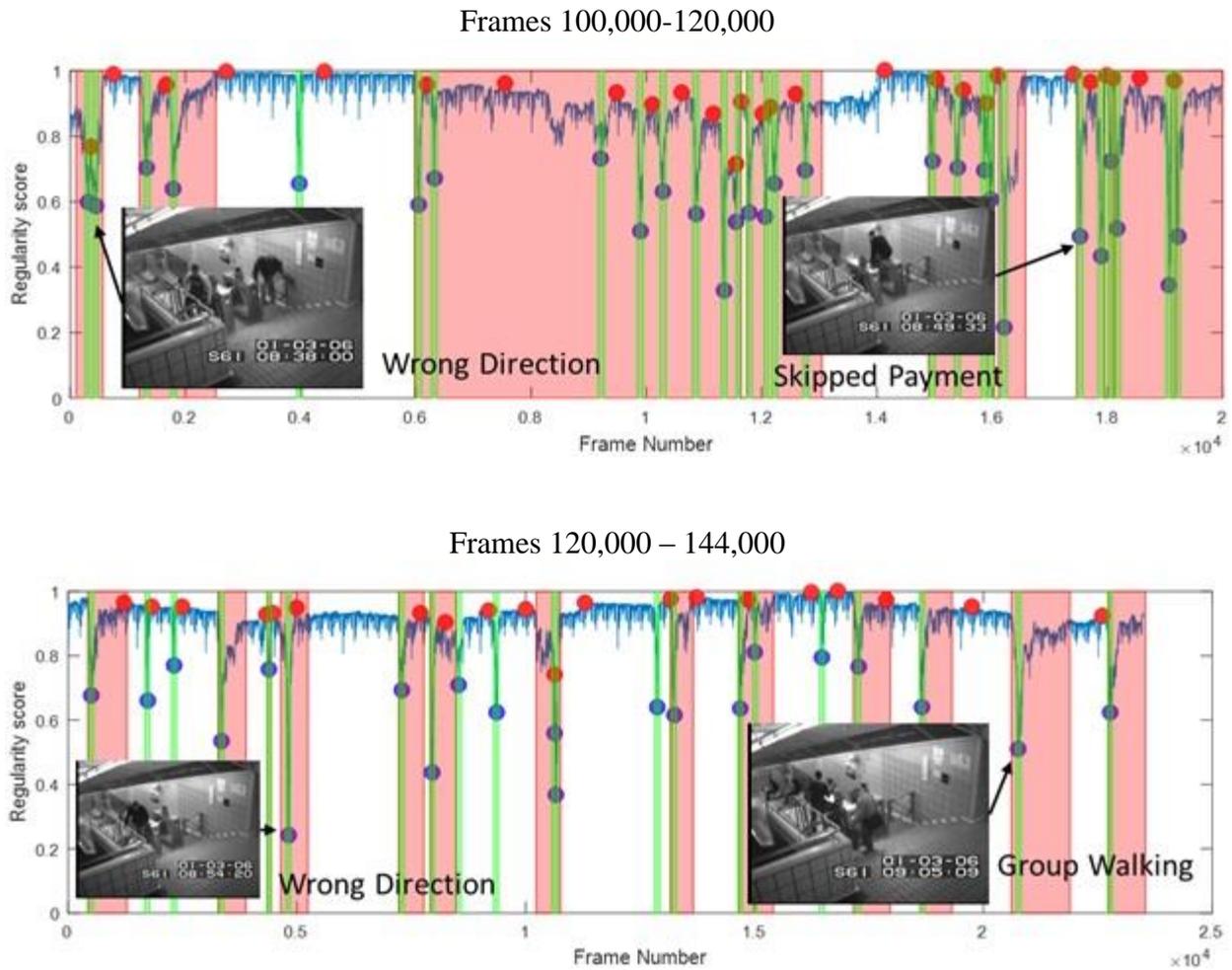

Frames 120,000 – 144,000

Figure 14. Anomaly evaluation graphs of test sequences from the Subway Entrance dataset.

## *2.4. Subway Exit*

As with the evaluation for the Subway Entrance, we use the 100 frame anomaly window despite longer target anomaly segments because there are many short anomalies in this video as well. Furthermore, it can be seen that the predicted anomalous segments still mostly occur within the target anomalous regions, allowing the anomalies to be found.

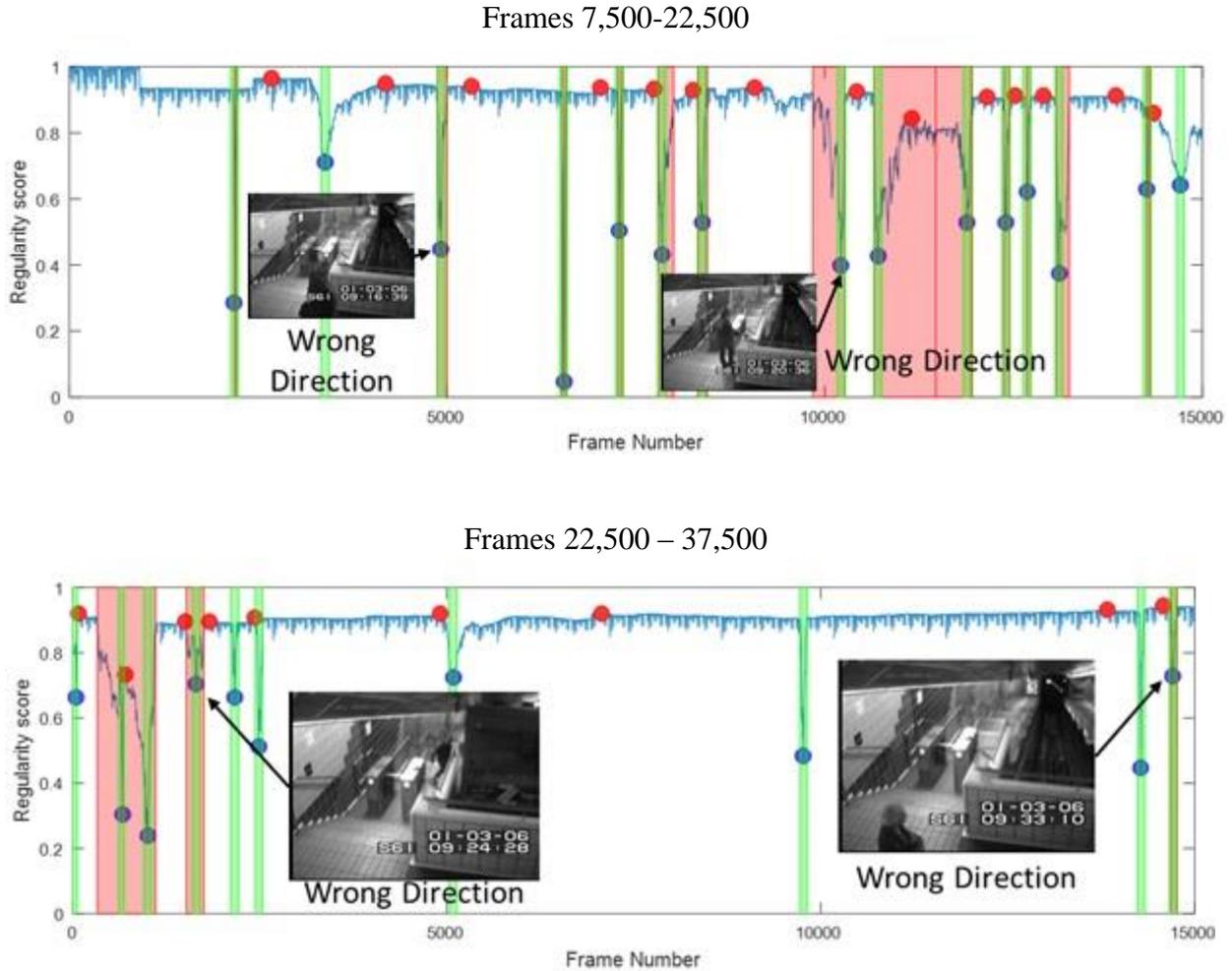

Figure 15. Anomaly evaluation graphs of test sequences from the Subway Exit dataset.

## *2.5. Avenue*

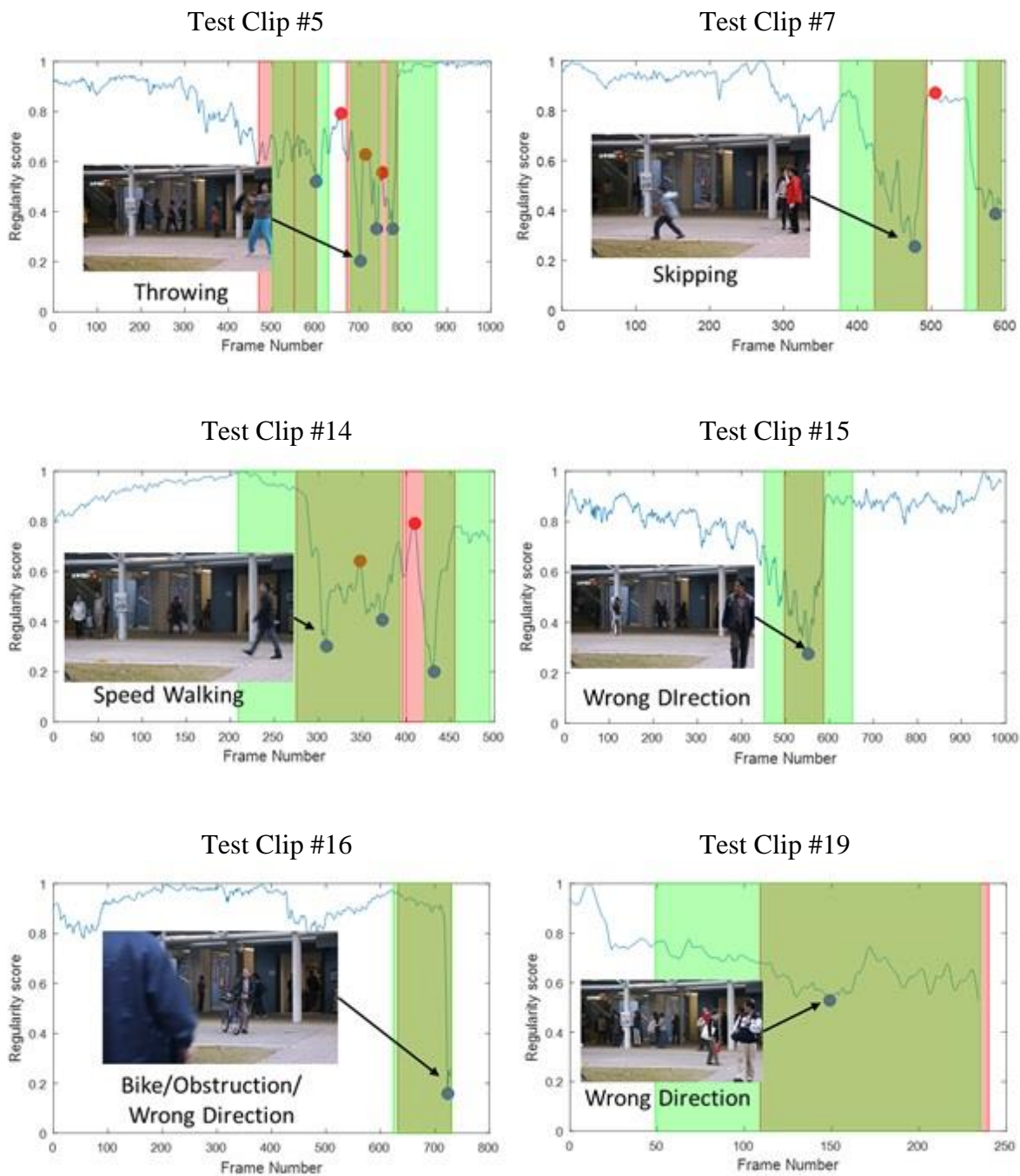

Figure 16. Anomaly evaluation graphs of test sequences from the Avenue dataset.